%% file: PAPER_main.tex
\documentclass[10pt,twocolumn,letterpaper]{article}

\usepackage[pagenumbers]{iccv} 
\input{preamble}

\definecolor{iccvblue}{rgb}{0.21,0.49,0.74}
\usepackage[pagebackref,breaklinks,colorlinks,allcolors=iccvblue]{hyperref}

\usepackage{tablefootnote}

\def\paperID{17} 
\def\confName{ICCV}
\def\confYear{2025}

\title{Linear Attention with Global Context: A Multipole Attention Mechanism for Vision and Physics}

\author{Alex Colagrande$^1$, Paul Caillon$^1$, Eva Feillet$^1$, Alexandre Allauzen$^{1,2}$ \\
\small{$^1$ Miles Team, LAMSADE, Université Paris Dauphine-PSL, Paris, France} \\
\small{$^2$ ESPCI PSL, Paris, France} \\
\small{\texttt{\{name\}.\{surname\}@dauphine.psl.eu}}
}

\begin{document}
\maketitle
\input{0_abstract}    
\input{1_intro}
\input{2_related_work}

\input{3_method}

\input{4_experiments}

\input{5_discussion}
\input{6_conclusion}
{
    \small
    \bibliographystyle{ieeenat_fullname}
    \bibliography{bibliography}
}
\clearpage
\newpage
\appendix

\normalsize
\input{appendix}

\end{document}

%% file: preamble.tex
\newcommand{\R}{\mathbb{R}}


%% file: 0_abstract.tex
\begin{abstract}
    Transformers have become the \textit{de facto} standard for a wide range of tasks, from image classification to physics simulations. Despite their impressive performance, the quadratic complexity of standard Transformers in both memory and time with respect to the input length makes them impractical for processing high-resolution inputs. Therefore, several variants have been proposed, the most successful relying on patchification, downsampling, or coarsening techniques — often at the cost of losing the finest-scale details. 
    In this work, we take a different approach. Inspired by state-of-the-art techniques in $n$-body numerical simulations, we cast attention as an interaction problem between grid points.
    We introduce the \textbf{Multipole Attention Neural Operator (MANO)} that computes attention in a distance-based multiscale fashion.
    MANO maintains, in each attention head, a \textbf{global receptive field} and has a \textbf{linear time and memory complexity} with respect to the number of grid points. Empirical results on image classification and Darcy flows demonstrate that MANO rivals state-of-the-art models, such as ViT and Swin transformer, while reducing runtime and peak memory usage by orders of magnitude. We open-source our code for reproducibility at: \url{https://github.com/AlexColagrande/MANO}.
\end{abstract}

%% file: 1_intro.tex
\section{Introduction}

Convolutional Neural Networks (CNNs) have formed the cornerstone of modern computer vision~\cite{lecun1989backpropagation,NIPS2012_alexnet,he2016deep}. Their architectural design leverages the spatial locality and translational invariance properties of images by applying shared convolutional filters over local receptive fields, enabling an efficient parameter usage and a strong inductive bias for grid-structured data. 

In recent years, Vision Transformers (ViTs)~\cite{vit_dosovitskiy2021} have emerged as an alternative to CNNs. They are based on the Transformer architecture~\cite{transformer_vaswani2017} introduced in the field of Natural Language Processing (NLP) for sequence-to-sequence learning. This neural architecture is characterized by the use of the self-attention mechanism~\cite{attention4NeuralMachineTranslation_bahdanau2016} that allows modeling global contextual information across the \textit{tokens} of a text or the \textit{patches} of an image. 
Despite lacking the strong locality priors of CNNs, attention-based architectures have demonstrated competitive performance in image classification, particularly when trained on large-scale datasets~\cite{oquab2023dinov2}.

Beyond computer vision and NLP, Transformer-based models have found application in scientific machine learning, particularly in the resolution of Partial Differential Equations (PDEs). 
PDEs constitute the fundamental mathematical framework for modeling a vast array of phenomena across the physical and life sciences - from molecular dynamics to fluid flows and climate evolution. 
Substantial efforts have been devoted to approximating the solution operators of such equations at scale.
Classical numerical solvers — including finite difference \cite{finite_differences}, finite element \cite{finite_elements}, and spectral methods \cite{spectral_methods}— discretize the underlying continuous operators, thereby recasting the problem as a finite-dimensional approximation. 
More recently, the increasing availability of observational data on structured grids has fostered a paradigm shift towards data-driven approaches such as Physics-Informed Neural Networks (PINNs) \cite{PINNS_raissi,deepxde,review_pinns}.
PINNs harness these observations to learn PDE solutions directly, enforcing physical consistency through soft constraints without relying on explicit mesh-based formulations. 
However, like classical numerical solvers, PINNs are typically designed to approximate the solution of a specific PDE instance — for example, computing the solution corresponding to a fixed coefficient, boundary or initial condition. 
This means even minor variations in input parameters require re-solving the system or, in the case of neural models, costly re-training. 
In contrast, operator learning \cite{survey_kovachki} targets a fundamentally more ambitious goal: to approximate a mapping between infinite-dimensional function spaces.

Although considerably more challenging, operator learning offers the advantage to generalize across input conditions without further optimization, offering a scalable and computationally efficient alternative to traditional point-wise solvers.
\textit{Note that operator learning is not restricted to PDEs, as images can naturally be viewed as real-valued functions on 2-dimensional domains}. 

As in computer vision, recent neural operator models benefit from the development of attention-based architectures~\cite{upt, Aurora}. 
However, attention suffers from quadratic time and memory complexity with respect to the input size, making it impractical for high-resolution data. 
To tackle this, some variants of the attention mechanism process data in local patches or down-sample the input, drastically cutting computational cost but often sacrificing the fine-grained details crucial for dense‐prediction tasks. 
Other methods replace full attention with low-rank approximations~\cite{wang2020linformer}, sparsity-inducing schemes~\cite{zaheer2020big}, or kernel-inspired formulations~\cite{choromanski2020rethinking}. 
Alternatively, Synthesizer proposes to learn attention weights without relying on explicit query–key products~\citep{tay2021synthesizer}. 
However, these approaches often trade off runtime for expressiveness or ease of implementation. 

In this work, we propose an efficient variant of the attention mechanism specifically suited for image classification as well as dense-prediction tasks such as physical simulations. 
Our method achieves computational gains by relaxing the classical attention formulation while preserving performance by preserving global context. 

We propose the \textbf{Multipole Attention Neural Operator} (MANO), a novel transformer neural operator 
in which each head computes attention between a point and a multiscale expansion of the input centered at that point.
The attention is performed against a hierarchical decomposition of the input, dynamically downsampled based on the query location.
Importantly, we compute the query, key and value matrices $Q, \ K$ and $V$ at every scale using the same point-wise operator to allow the model to accept inputs at any resolution.

\begin{figure*}
    \centering
    \includegraphics[width=\linewidth]{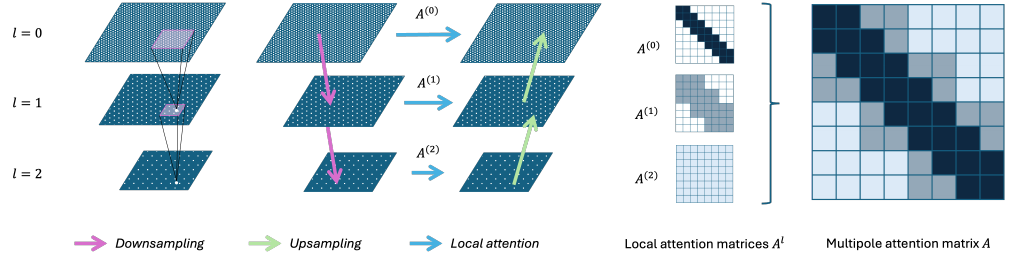}
    \caption{(Left) The multi-scale grid structure. (Center) The V-cycle structure for computing multipole attention with the fast multipole method. (Right) Attention matrices. Illustration with three levels. The attention matrix $A$ is computed in a multiscale manner with respect to each level. The higher the level, the shorter the range of the interaction. At a given layer, down-sampling (resp.  up-sampling) is performed using a convolution kernel (resp. deconvolution) shared across all different levels.
}
    \label{fig:method}
\end{figure*}

\textbf{Our contributions are as follows:}
\begin{itemize}
    \item We propose the Multipole Attention Neural Operator (MANO) that formulates attention as an interaction problem and solves it using the Fast Multipole Method.
    \item By combining MANO with the Swinv2 architecture, we improve transfer learning results on several image classification tasks. 

    \item MANO achieves state-of-the art results on Darcy flow simulation benchmarks matching, and sometimes surpassing, state of the art baselines. 
\end{itemize}

%% file: 2_related_work.tex
\section{Related work}
The Vision Transformer (ViT) \cite{vit_dosovitskiy2021} was the first to successfully adapt the Transformer architecture to image classification, achieving remarkable performance. It divides the input image into fixed-size patches, flattens them into token embeddings, adds positional encodings, and processes the resulting sequence with a Transformer encoder. When pretrained on large datasets such as ImageNet-21k~\cite{imagenet} or LVD-142M\cite{oquab2023dinov2}, 
ViTs rival or exceed CNNs on image classification tasks. 
However, despite their efficiency, they suffer from limited local information interaction, single-feature representation and therefore low-resolution outputs making it sub-optimal for dense prediction tasks.

These limitations have motivated a number of efficient vision transformer variants.

\subsection{Efficient Vision Transformer Variants}

\paragraph{Swin Transformers}
The Swin Transformer \cite{liu2021swin} restricts self‐attention to non‐overlapping windows that are shifted between layers, yielding hierarchical, multi‐scale representations without global attention.  
Swin Transformer V2 \cite{liu2022swinV2} augments this design with learnable‐temperature scaled cosine attention, log‐spaced relative position bias, and continuous pre-norm, improving high‐resolution stability and enabling deeper networks—all while preserving the original’s efficient window‐based computation.

\paragraph{Distilled and Compact ViTs}:  
TinyViT \cite{wu2022tinyvit} uses \emph{pretraining-stage distillation} from a large teacher (e.g., Swin-B/L trained on ImageNet-21k). By caching teacher logits and applying neural architecture search under FLOPs/parameter constraints, TinyViT produces smaller models at only a small performance loss. 

Data-Efficient Image Transformers (DeiT) \cite{touvron2021going} add a learnable \emph{distillation token} that learns from a CNN teacher’s soft logits.  
Later work~\cite{touvron2022deit} adds self-supervised distillation and token pruning for further efficiency.

Collectively, these efforts have greatly extended ViT applicability across resource-constrained tasks. However, the inherent multi-scale structure of images remains only partially integrated into existing alternatives to the attention mechanisms, potentially hindering the overall performance.

\subsection{Operator Learning via Multipole Attention}
  
In this work, we illustrate the interest of our proposed multipole attention mechanism for learning solution operators of PDEs directly from input–output pairs, as encountered in tasks like fluid flow estimation and other dense prediction problems \cite{survey_kovachki}. Operator learning was first explored by \citet{lu2019deeponet}, who established a universal approximation theorem for nonlinear operators using DeepONets, laying theoretical foundations for neural operator approximation. Building on this foundation, the Fourier Neural Operator (FNO) \citep{FNO} parameterizes an integral kernel in the Fourier domain—using efficient FFT-based convolutions to capture global interactions across the entire domain. These pioneering methods have since inspired a wealth of extensions—but their reliance on global or Fourier‐based interactions limits their scalability to very high-resolution grids.
These works paved the way for numerous extensions.

\paragraph{Transformer neural operators} In \cite{fourier_galerkin_transformer} the classical transformer 
was adapted for the first time to operator learning problems related to PDEs. The paper explores two variants, respectively based on the Fourier transform and on the Galerkin method. 
The latter one uses a simplified attention based operator, without softmax normalization. This solutions 
shares the linear complexity with our work but not the same expressivity. 
In this line of work, LOCA \cite{coupled_attention} uses kernel theory to map the
input functions to a finite set of features and attends to them by output query location. 
Recently, \citep{calvello2024continuum} proposed to handle attention in a continuous setting and, as well as \cite{wang2024cvit}, proposed an operator-learning version of the more classical ViT.
Notably, the Universal Physics Transformer (UPT)~\cite{upt} scales efficiently based on a coarsening of the input mesh.

\paragraph{Multiscale numerical solvers.} 
Our method is inspired by multi-scale numerical solvers~\citep{brandt1977multi,briggs2000multigrid,hackbusch2013multi}, in particular the Fast Multipole Method (FMM). A new version of the Fast Multipole Method is introduced by \cite{FMM} for the evaluation of potential fields in three dimensions and its specialization with the V-cycle algorithm, introduced by \citep{FMM_V_cycle}.

\paragraph{Theoretical studies on transformers.}
The particle-interaction interpretations of attention was first introduced in Sinkformer~\cite{sinkformer}. \cite{cluster_particles_attention} views Transformers as interacting particle systems, they described the geometry of learned representations when the weights are not time dependent, and \cite{math_attention} developed a mathematical framework for analyzing Transformers based on their interpretation as interacting particle systems inspiring us to compute the attention using the most efficient techniques available for solving particle-interaction problems. 

\paragraph{Multiscale neural architectures.}
Several transformer architectures related to the multiscale principle used in our method were proposed in the one-dimensional setting of Natural Language Processing (NLP)~\cite{H-transformer1D,MRA-attention,FMMformer,fastmultipoleattention}, and in graph learning methods ~\cite{mgno, migus2022multi, zhdanov2025erwin}

\paragraph{Relation to Fast Multipole Attention}
Among existing approaches, the closest to ours is Fast Multipole Attention (FMA) \cite{fastmultipoleattention}, which reduces the $O(N^2)$ cost of 1D self‐attention via hierarchical grouping: nearby queries attend at full resolution, while distant keys are merged into low‐rank summaries, achieving $O(N\log N)$ or $O(N)$.  

Our method differs in two key aspects:
\begin{itemize}
  \item \textbf{Input domain.} FMA targets one‐dimensional token sequences; we operate on two‐dimensional image grids with multiscale spatial windows.
  \item \textbf{Downsampling.} FMA hierarchically downsamples queries, keys, and values. In contrast, we downsample the input feature map \emph{prior} to attention, yielding a self‐contained block that integrates seamlessly with standard transformer backbones (e.g., SwinV2) and preserves pretrained attention weights.
\end{itemize}

%% file: 3_method.tex
\begin{figure*}[htbp]
    \centering
      \includegraphics[width=1\textwidth]{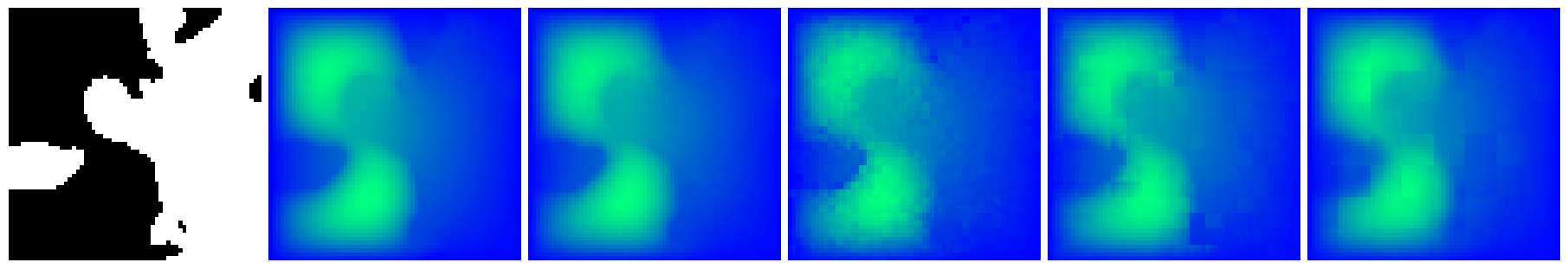}
        \caption{Darcy flow reconstruction: from left to right—input coefficient field, ground truth solution, MANO prediction, and ViT predictions using patch sizes 2, 4, and 8. MANO applies multipole attention using overlapping windows of size $2$, and performs downsampling and upsampling across $5$ levels using convolutions with kernel size $2 \times 2$, stride $2$, and zero padding.}
    \label{fig:mano_ill}
\end{figure*}

\section{Introducing MANO}

\subsection{Attention as an interaction problem} \label{subsec:Attention_as_an_interaction_problem} 
In this section, we cast the computation of self-attention as a dense $n$-body interaction problem. An $n$-body system consists of $n$ entities (often referred to as bodies) whose state is described by a configuration $(x_1, \ldots, x_n)$. The evolution of such a system is governed by a set of interaction laws, which in our setting are determined by pairwise interactions specified through a kernel function:
\begin{align*}
\kappa: \mathbb{R}^d \times \mathbb{R}^d &\rightarrow \mathbb{R} \\
(x_i, x_j) &\mapsto \kappa(x_i, x_j)
\end{align*}
An $n$-body simulation refers to a numerical method for computing these interactions, typically requiring $O(n^2)$ operations due to the dense pairwise structure. This computational cost motivates the development of faster approximations, such as the Fast Multipole Method, which reduces the complexity to $O(n)$.
\\
In the following, we consider a regular grid, corresponding either to the pixels of an image in the classification setting or to the discretized samples of an input function in the operator learning framework.
We denote by $X$ a sequence of $N$ observations $(x_1, \ldots, x_N)^\top \in \R^{N \times d}$ with elements embedded in dimension $d$. 
The self-attention mechanism firstly applies three learnable linear projections to obtain queries, keys and values~\cite{transformer_vaswani2017}:
\begin{equation}
    Q = X W_q, \quad K =X W_k, \quad V = X W_v
\end{equation}
with $W_q, W_k, W_v \in \R^{d \times d}$ and $b_q, b_k, b_v \in \R^d$.
Next, it computes an $N\times N$ attention matrix $A$ whose $i$-th row forms a probability distribution over all keys:
\begin{equation}
  A_{ij} = \frac{\exp\!\bigl(Q_i^\top K_j / \sqrt{d}\bigr)}
         {\displaystyle\sum_{l=1}^N \exp\!\bigl(Q_i^\top K_l / \sqrt{d}\bigr)}.
\end{equation}
Finally, each token is updated as a convex combination of the value vectors:
\(
    x_i \leftarrow \sum_{i=j}^n A_{i, j}V_j, 
\)
In this form, one can view the set $\{x_i\}_{i=1}^N$ as a cloud of $N$ particles in $\R^d$ interacting in a pairwise manner via a kernel $\kappa$ defined as:
\begin{equation}
    \kappa(Q_i,K_j) = \exp\!\bigl(Q_i^\top K_j / \sqrt{d}\bigr).
\end{equation}

Therefore, we interpret a self‐attention layer as a single time step of a discretized $N$-body dynamical system. Under this analogy, computing attention is equivalent to predicting the next state of an interacting particle system — and it becomes natural to accelerate this computation using the FMM~\cite{FMM}, reducing the usual $O(N^2)$ cost of the pairwise sums to $O(N)$.

\subsection{MANO}
In this section we detail the Multipole Attention layer as well as the complexity of the model.

\paragraph{Method Overview}:
Let $X_0 = X\in \mathbb{R}^{H\times W\times d}$ be the original high‐resolution image (height $H$, width $W$, embedding dimension $d$). We define $L$ levels of downsampling by a convolutional kernel $D$ with weights shared accross levels, producing
$X_\ell = D \bigl(X_{\ell-1}\bigr)\quad(\ell=1,\dots,L)$
where $X_\ell\in\mathbb{R}^{H/2^\ell \times W/2^\ell \times d}$. At each level $\ell$, we partition the feature map into potentially overlapping sliding windows and, within each window, compute the attention map
$A_\ell = \mathrm{Softmax}\bigl(Q_\ell K_\ell^\top / \sqrt{d} \bigr)$
where \(Q_\ell,K_\ell,V_\ell\in\mathbb{R}^{(H/2^\ell \times W/2^\ell)\times d}\) are the query, key, and value embeddings extracted from \(X_\ell\). 
This restricts self-attention to localized neighborhoods while still enabling cross-window interactions via the sliding overlap
and the hierarchical mixing. 

We then produce the attended features $\tilde X_\ell = A_\ell \,V_\ell$,
and upsample back to the next‐finer resolution via a transposed convolution $U$:
\begin{equation}
\hat X_\ell = U\bigl(\tilde X_\ell\bigr) \in \mathbb{R}^{H/2^{\ell-1}\times W/2^{\ell-1}\times d}.
\end{equation}
Finally, we combine all levels by summation at the original resolution: 
$X_{\text{out}} = \sum_{\ell=0}^L U^ \ell \bigl(\,\mathrm{Attn}\bigl(X_\ell\bigr)\bigr)$,
where $\mathrm{Attn}(X_\ell)$ denotes $\tilde X_\ell$ at level $\ell$, and $U^0$ is the identity.

Sharing the same convolutional kernel for both down-sampling and up-sampling—and reusing the same attention weights—keeps the total parameter count constant, regardless of the number of layers $L$. The convolutions have the role to provide a representation of the input at the next scale, independently of the scale. This ensures that an attention map learned at the finest scale produce effective representations at different scale, even in the case of a pretrained attention from a windowed-attention based model, such as the SwinV2, with finetuning on the convolutional paramters but not on the attention weights. The shared convolutions act as scale-agnostic projectors, producing the next-scale feature representation in the same way at every level. As a result, an attention map learned at the finest resolution remains effective across all scales. In practice, this means one can take a windowed-attention backbone, such as SwinV2, freeze its attention weights, and fine-tune only the convolutional parameters to adapt it to new resolutions without increasing model size.

\paragraph{Computational Complexity.}
We analyze the cost under non‐overlapping windows on a square image of side $H$ (so $N=H^2$ tokens), embedding dimension $d$, window size $w$, and down‐sampling factor $k$. The maximum number of levels is $L=\log_k(H)$. We denote by $N_\ell = \frac{N}{k^{2\ell}}$ the number of grid points at level $\ell$.

Each windowed self‐attention on $M = w^2$ tokens costs $O(M^2 d)$ and is applied across $O(N_\ell / M)$ windows, for a total complexity of
\begin{equation}
    O(N_\ell M d)
\end{equation}
Windowed attention computes a standard self-attention for a complexity of $O(M^2 d)$ whithin each of the $N/M$ windows. The total complexity is thus $O(NMd)$. Our Multipole attention iterates the same windowed attention and aggregates the contribution at each level $\ell$ therefore the total complexity reads
\begin{equation}
    \sum_{\ell = 0}^{L} N_\ell M d = \sum_{l = 0}^{L} O\left(\frac{N}{k^{2 \ell}} M d\right) = O(N M d)
\end{equation}
So interestingly, even though we apply windowed attention across multiple scales, the total cost remains dominated by the finest‐scale pass, with coarser levels adding a negligible additional overhead. As a result, our approach preserves the linear complexity of single‐scale windowed attention while delivering significantly greater expressive power.

%% file: 4_experiments.tex
\begin{table*}[htbp]
    \centering 
    \resizebox{1.8\columnwidth}{!}{%
    \begin{tabular}{lllllllll}
      \toprule
      Model        & Params. &   Complexity   & Tiny-IN-202             & Cifar-100       & Flowers-102       & Food-101        & StanfordCars-196 & OxfordIIITPet\\
      \midrule
      TinyViT~\cite{wu2022tinyvit}          & $21$M &  $O(N^2)$                             &  -                & $75.2\%$        & $82.4\%$          &  -               & $\underline{61.7 \%}$ & $86.5 \%$\\
      ViT-base \cite{vit_dosovitskiy2021}          &  $86$M & $O(N^2)$  & $73,07\%$ & $\underline{80.63 \%}$ & $\mathbf{92.75 \%}$ & $\underline{80.31\%}$ &  $41.95\%$ & $\underline{87.68 \%}$\\
      DeiT-small \cite{touvron2022deit}       & $22$M & $O(N^2)$  &  $\underline{81.34\%}$ & $75.15 \%$       &   $66.60\%$ & $71.39\%$ & $36.38 \%$   & $87.56 \%$ \\
      SwinV2-T \cite{liu2022swinV2}     &  $28$M  & $O(N)$        &    $80.53\%$                 & $75.47\%$     &           $56.46\%$                &      $76.96\%$            & $38.36\%$ & $87.14\%$\\
      \midrule
      MANO-tiny     &   $28$M &  $O(N)$       &   $\mathbf{87.52 \%}$   &  $\mathbf{85.08\%}$  & $\underline{89.00\%}$ & $\mathbf{82.48\%}$  & $\mathbf{65.68 \%}$ & $\mathbf{88.31\%}$\\
      \bottomrule
    \end{tabular}
    }
    \caption{Linear probing accuracies for several image classification datasets. MANO matches or even outperforms the performances of state-of-the-art models. The results for TinyViT are taken from \cite{wu2022tinyvit}, while all other baselines are fine-tuned via linear probing using pretrained backbones\tablefootnote{Checkpoints are available at \url{https://huggingface.co/google/vit-base-patch16-224} (ViT-base), \url{https://huggingface.co/facebook/deit-small-patch16-224} (DeiT-Small), and \url{https://huggingface.co/timm/swinv2_tiny_window8_256.ms_in1k} (SwinV2-T).}.
    For each model, we report the number of parameters and the asymptotic complexity of its attention block with respect to the number of patches $N$. As a plug-and-play replacement for the attention mechanism, our attention can be applied after patching, similar to Swin, allowing our vision-specific MANO variant to scale linearly with the number of patches. } 
        \label{tab:img_res}
\end{table*}

\section{Experimental settings}
This section outlines the experimental setup for both image classification and physics simulations. 

For image classification, we evaluate on several fine-grained datasets a Swin Transformer V2 modified with our proposed attention mechanism. 
Models are initialized with weights pretrained on ImageNet-1k. The encoder is frozen (except for the additional convolutions of MANO) and a linear classifier is learnt on the target classification tasks. 

For physics simulations, we train all models on instances of the Darcy flow problem, from scratch and across different resolutions.

\subsection{Image classification}

\paragraph{Datasets.} The ImageNet-1k~\cite{imagenet} dataset is used to pretrain all models and we perform \emph{linear probing} on several downstream classification benchmarks, namely CIFAR-100\cite{krizhevsky2009learning}, Oxford Flowers-102~\cite{nilsback2008automated}, Stanford Cars~\cite{krause20133d}, Food101~\cite{bossard2014food}, Tiny-ImageNet-202~\cite{le2015tiny} and Oxford-IIIT Pet Dataset~\cite{parkhi2012cats}. 

\paragraph{Architecture.} 
As the backbone for our Multipole Attention model, we adopt the ``Tiny'' version of the Swin Transformer V2~\citep{liu2022swinV2}.
Since the attention block is shared across all levels of the multipole hierarchy, MANO can inherit the pretrained weights from the original Swin Transformer, requiring only a small number of additional trainable parameters: one convolution and one transposed convolution per attention head, along with the classification head. Convolutions have a kernel size of 2 and a stride of 2. 
This design ensures that our variant introduces a minimal increase in parameter count relative to the base model. Specifically, for the Tiny version, the total number of parameters increases from $~27,73$M to $~28,47$M, corresponding to an additional $740,356$ parameters, or $~2.67\%$ more than the original model. 

\paragraph{Training.}
We freeze the pretrained encoder weights and train a single fully connected layer for $50$ epochs on top of the frozen encoder using AdamW as optimizer with a cosine annealing learning rate schedule. 
Training the models with the original shifted attention of Swinv2 and the models with our proposed multipole attention only differs by a warm-up phase to learn the upsampling and downsampling convolutional filters introduced by MANO. 
We report the resulting top-1 accuracy of these experiments in Table~\ref{tab:img_res}.

\subsection{Darcy flow simulation} 

\paragraph{Task}
We evaluate our method on the task of steady-state 2D Darcy Flow simulation, a widely used task in the neural operator literature~\cite{survey_kovachki}. 
The problem is based on the following second-order, linear elliptic PDE:
\begin{equation}
-\nabla \cdot (a(x) \nabla u(x)) = f(x), \quad x \in (0,1)^2,
\end{equation}
to solve with homogeneous Dirichlet boundary conditions:
\(
u(x) = 0, \quad x \in \partial(0,1)^2.
\)
In this PDE, the function $a(x)$ represents the spatially varying permeability of a porous medium. 
The forcing term $f(x)$ is fixed to $f(x) \equiv 1$ across all inputs. The output $u(x)$ is the scalar field representing the pressure within the domain. Although the PDE is linear in $u$, the map from the input $a(x)$ to the solution $u(x)$ is nonlinear due to the interaction of $a(x)$ with the gradient operator inside the divergence.

The task is to learn this solution operator: given a new input field $a(x)$, the model must predict the corresponding output $u(x)$. In our experiments, $a(x)$ is sampled as a binary field (i.e., values are either 0 or 1), representing a medium composed of two different materials. 

\paragraph{Architecture.} We use a classical transformer architecture of depth $8$ with $4$ attention heads per layer. In place of the conventional self-attention, we employ our multipole attention module. Additionally, we apply Layer Normalization at every level to improve training stability and mitigate issues of vanishing or exploding gradients, which can arise due to shared attention across hierarchical levels.

\paragraph{Training.} We train all the considered models for $50$ epochs with AdamW optimizer and cosine learning rate scheduler. The initial learning rate is in the order of $10^{-4}$.
We use the dataset open-sourced in~\cite{survey_kovachki}, comprised of input–output pairs $(a, u)$ at resolutions $n \times n$ for $n \in \{16, 32, 64 \}$. The model is trained to minimize the mean squared error (MSE) on the training set and evaluated on a held-out test set using the relative MSE error, where $\hat{u}$ is the model prediction and $u$ the ground truth solution:
\(
\frac{| \hat{u} - u |_2}{| u |_2}.
\)

\section{Results}

\subsection{Image Classification Results}

Table~\ref{tab:img_res} presents the top-1 accuracy of three models of similar parameter counts (ViT
($21$M parameters), SwinV2-T ($28$M), ViT-base ($86$M), DeiT-small ($22$M) and MANO ($28$M)) across six downstream image classification datasets.
The reported results for TinyViT are taken from \cite{chen2021vision,wu2022tinyvit} while the results for SwinV2-T are taken from \cite{liu2022swinV2}.

First, MANO consistently outperforms TinyViT on all benchmarks where results are available. It also surpasses DeiT-small and SwinV2-T when these models are fine-tuned via linear probing using ImageNet-1k pretrained weights, across an expanded set of datasets. Compared with the bigger ViT-base, our model performs better in all the benchmark except than Flowers-$102$ that is the dataset with less training data among the one considered. the dataset with the smallest training set among those considered. This suggests that in low-data regimes, models with a higher parameter count may have an advantage due to their increased capacity to memorize or adapt to limited supervision.

The improvement ranges from about 1–2 \% on easier tasks like Oxford–IIIT Pet to nearly $5/7$ points on more challenging datasets such as CIFAR–$100$ and Tiny-ImageNet. Even compared to ViT-base, which has more than twice the number of parameters, MANO achieves gains of roughly 3–10 points accross all the benchmarks expect for Flowers-$102$, demonstrating that multiscale hierarchical attention produces significantly more transferable features without increasing too much model size.

Second, the advantage of MANO becomes especially pronounced on fine-grained classification tasks. On Flowers–102 and Stanford Cars, SwinV2-T achieves only $56.5 \%$ and $38.4 \%$ accuracy, respectively, while TinyViT recovers to $82.4 \%$ and $61.7 \%$. In both cases, MANO further improves performance to $89.0 \%$ on Flowers–$102$ and $65.7 \%$ on Cars, indicating that combining local details (e.g., petal shapes or headlight contours) with global context (overall flower appearance or car silhouette) is critical for distinguishing highly similar classes.

Third, on medium-difficulty datasets such as Tiny–ImageNet–$202$ and CIFAR–$100$, MANO again holds a clear lead. It outperforms SwinV2-T by approximately $7$ points on Tiny–ImageNet–$202$ and by around $10$ points on CIFAR–$100$. These results suggest that attending to multiple resolutions—capturing both fine textures and broader scene structures—yields better representations than the single-scale windowed attention used in SwinV2-T.

Finally, although MANO and SwinV2-T share the same parameter count (28M), MANO delivers a consistent 5–10 point advantage on mid-level benchmarks and maintains a smaller lead on easier tasks like Oxford–IIIT Pet. TinyViT’s 21M parameters are insufficient to match either 28M model, underscoring that hierarchical multiscale attention makes more effective use of model capacity than either pure global self-attention (ViT) or fixed-window local attention (SwinV2-T).

\subsection{Darcy Flow Simulation Results}
\begin{table}[htbp]
    \centering
    \label{tab: darcy_res}
    \resizebox{0.8\columnwidth}{!}{
    \begin{tabular}{llll}
      \toprule
      Model & $16 \times 16$ & $32 \times 32$ & $64 \times 64$ \\
      \midrule
      FNO                 &   $0.0195$ &  $0.0050$    &  $0.0035$  \\
      ViT  patch\_size=8  &   $0.0160$ &  $\underline{0.0038}$    &  $0.0021$           \\
      ViT  patch\_size=4  &   $0.0179$ &  $0.0039$    &  $\underline{0.0019}$         \\
      ViT  patch\_size=2  &   $0.0169$ &  $0.0049$    &  $0.0026$       \\
      Local Attention     &   $\underline{0.0133}$ &  $0.0188$    &  $0.0431$      \\
      MANO                &   $\mathbf{0.0080}$ &  $\mathbf{0.0020}$    &  $\mathbf{0.0013}$      \\
      \bottomrule
    \end{tabular}
    }
    \caption{Benchmark on Darcy flow simulations. Relative MSE. A given model is evaluated and tested on the same resolution, either $16 \times 16$, $32 \times 32$ or $64 \times 64$.}
    \label{tab:mano_darcy}
\end{table}

Table~\ref{tab:mano_darcy} presents the relative MSE of various models trained from scratch on Darcy flow at different resolutions. 
The Fourier Neural Operator (FNO) \cite{FNO} is a neural operator designed to learn mappings between function spaces, such as the coefficient-to-solution map for PDEs. By operating in the Fourier domain, the FNO captures long-range dependencies across the entire domain with near-linear complexity \(O(N^2 \log N)\) (for an \(N\times N\) grid), making it effective for a wide range of PDE-based tasks, leading to state-of-the-art results on tasks such as Darcy Flow Simulation.
It achieves MSEs of 0.0195, 0.0050, and 0.0035 as the grid is refined, showcasing its strength in capturing global spectral components but its limited ability to resolve fine-scale details at coarser resolutions. 

For ViT, we evaluate patch sizes of $8$, $4$, and $2$: the patch-$4$ variant attains the best errors ($0.0179$, $0.0039$, $0.0019$), whereas smaller patches (size $2$) slightly worsen performance at low resolution and fail to match patch-$4$ at $64^2$. This sensitivity indicates that vanilla ViT’s pure global attention is capable of approximating the solution operator but depends heavily on patch granularity. 
In contrast, a pure local-attention model (fixed window) degrades sharply at $32^2$ and $64^2$, since local windows cannot propagate long-range dependencies across the domain. 

By combining fine-grid attention (to capture local conductivity channels) with progressively coarser resolutions (to model global pressure fields), MANO consistently achieves the lowest errors, roughly halving the MSE of both FNO and standard ViT at every scale and overcoming the locality limitations inherent in fixed-window attention.

Lastly, we examine in Figure~\ref{fig:mano_ill} the quality of the reconstructed images, demonstrating that MANO’s multiscale modeling recovers both sharp transitions and smooth boundaries with high fidelity, even on a coarse grid. By contrast, images reconstructed by ViT exhibit noticeable patching artifacts.

In summary, MANO’s multiscale hierarchical attention achieves state-of-the-art performance on both Darcy flow simulations and image classification tasks. Its design makes it well suited to the corresponding data, as it captures fine‐scale detail and broad‐scale context simultaneously.  

\begin{figure}[htp]
    \centering
      \includegraphics[width=\linewidth]{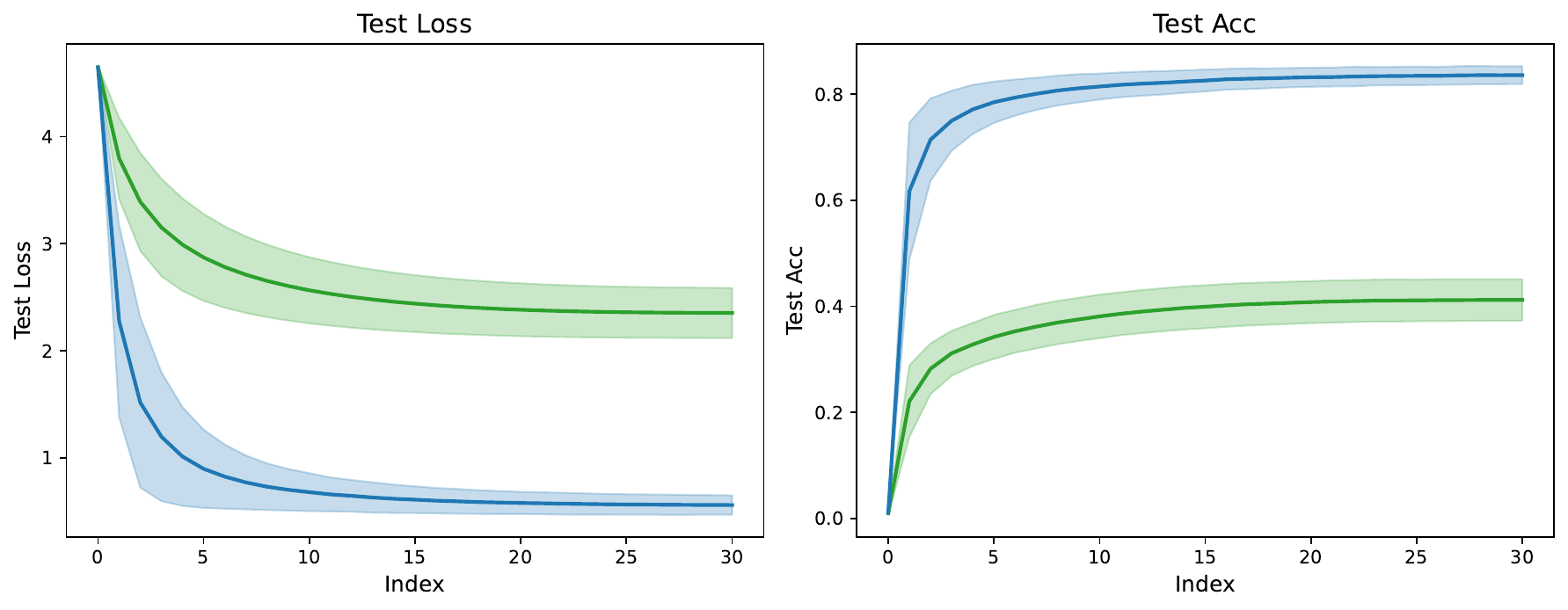}
        \caption{Ablation study comparing average pooling (green) and learnable convolutions (blue) for the sampling step in MANO. We report the Cross-Entropy validation loss (left) and accuracy (right) on CIFAR-100. Mean and standard deviation over fifteen runs are reported, varying the learning rate between $10^{-3}$ and $10^{-4}$.}
    \label{fig:Ablation_pool_conv}
\end{figure}

\paragraph{Hyperparameters.} 
A detailed table of hyperparameters is provided in the Appendix; below, we outline our main design choices. 

The “number of levels” specifies how many hierarchical scales are included in the multipole attention; with a window size of 8 for the windowed attention, we achieve the best performance using the maximum of 3 levels. Due to Swin’s built‐in downsampling between stages, this corresponds to 3 levels across the first two layers, 2 levels for the next two layers, and a single level for the remaining eight layers.  
To coarsen the input grid, we compared average pooling versus learned convolutions—convolutions consistently outperformed pooling. For upsampling, transpose convolutions outperformed nearest‐neighbor interpolation. When used, \texttt{kernel\_size} and \texttt{stride} refer to these convolutional operations. 

As shown in Figure~\ref{fig:Ablation_pool_conv}, using learned convolutions for both down- and up-sampling significantly improves expressivity: even with pretrained attention weights, convolution-based sampling enables a windowed attention trained at one resolution to transfer effectively to another. Note that a single convolutional kernel is reused for all downsampling operations, and a separate kernel is reused for all upsampling operations.

%% file: 5_discussion.tex
\section{Discussion}

\paragraph{Hierarchy depth in vision vs. Physics.}
In our image classification experiments, we follow SwinV2-T’s architecture and compute attention at three hierarchical levels in early stages, two in the middle, one at the end of the encoder). For Darcy flow grids, we set downsampling steps so that the coarsest scale is $2 \times 2$, the smallest possible when using a $2 \times 2$ window for the attenion. We found that increasing the number of levels consistently improves performance, however we note that in physics simulation, the number of levels can be treated as a hyperparameter and tuned based on input resolution and the desired balance between local and global interactions.

\paragraph{Limitations.} 
MANO’s current design uses a fixed, static hierarchy and attention parametrization. While effective, it could benefit from learnable scale selection and explicit cross‐level interactions to better capture multi-scale couplings. 
Additionally, we assume a uniform grid to discretize the input. This simplifies implementation but fails to capture regions with steep gradients or intricate boundaries in physics simulations. 
Introducing adaptive meshing would thus improve accuracy and efficiency for localized phenomena that otherwise demand a higher resolution simulation, but would require redefining attention over nonuniform spatial supports. 
Finally, our current implementation could be accelerated by a hardware—optimized GPU kernel to further reduce runtime.

\paragraph{Future Directions.}
Beyond steady‐state Darcy flow, MANO can naturally extend to time‐dependent PDEs by integrating with recurrent timestepping or operator‐splitting schemes, preserving its ability to capture both spatial multiscale structure and temporal evolution. We also plan to extend our method to unstructured meshes and irregular domains common in real‐world physics simulations. 
On the computer‐vision side, applying MANO to dense prediction tasks—such as semantic segmentation or image inpainting—is promising, since its multiscale attention could dynamically balance local details and global context more effectively than standard U-Net architectures.  

%% file: 6_conclusion.tex
\section{Conclusion}

We propose \textbf{MANO}, an efficient attention‐based architecture inspired by \(n\)‐body methods, which interprets attention as interactions among mesh points. By introducing a distance‐based, multiscale attention mechanism, MANO achieves linear time and memory complexity per head while preserving a global receptive field. Across several image classification benchmarks and Darcy flow simulations, MANO matches the accuracy of full‐attention models yet substantially reduces runtime and peak memory. Unlike patch‐based approximations, it avoids discontinuities and retains long‐range dependencies inherent to physical systems. Our results demonstrate that MANO is a scalable alternative for both vision tasks and mesh‐based simulations. Future work includes applying MANO to dense vision tasks such as semantic segmentation, and to extend it to irregular meshes and a broader class of physical simulations.  

\section*{Acknowledgments}
This work was supported by the SHARP ANR project ANR-23-PEIA-0008 in the
context of the France 2030 program and by the HPC resources from GENCI-IDRIS (grants AD011015154 and A0151014627).

%% file: appendix.tex
\section{Extended related work section}
In this section, we provide a more comprehensive overview of the related literature, expanding upon the works briefly mentioned in the main text. 

\paragraph{Vision Transformers (ViTs)}
(ViTs)~\cite{vit_dosovitskiy2021} divide each input image into fixed-size patches (e.g., $16 \times 16$), flatten them into tokens, add positional embeddings, and process the resulting sequence with a Transformer encoder. When pretrained on large-scale datasets such as ImageNet-21k~\cite{imagenet} or LVD-142M~\cite{oquab2023dinov2}, ViTs achieve performance on par with or surpassing that of convolutional neural networks (CNNs) on standard image classification benchmarks.

Despite these advances, ViTs face several limitations: 
\begin{enumerate}
    \item quadratic computational complexity $\mathcal{O}(N^2)$ with respect to the number of input patches ($N$, where typically $N \approx 196$ for a $224 \times 224$ image).
    \item Absence of built-in locality and translation equivariance, in contrast to CNNs, which makes ViTs more dependent on large training datasets.
    \item High computational and memory demands—for instance, ViT-Large/$16$ contains roughly $300$ million parameters and requires thousands of GPU-hours to train~\cite{vit_dosovitskiy2021}.
\end{enumerate}
These drawbacks have spurred the development of more efficient ViT variants.

\subsection{Efficient Vision Transformer Variants}
Several efficient alternatives to the standard attention have been proposed in the literature to address the limitations of Vision Transformers. While they differ in methodology, they have collectively inspired this work.
\paragraph{Linear-Attention Transformers:}   
Linformer \cite{wang2020linformer} projects keys and values into a low-dimensional subspace (\(k\ll N\)), reducing per-head complexity from \(O(N^2)\) to \(O(Nk)\) while retaining competitive accuracy. Performer \cite{choromanski2020rethinking} uses a randomized feature map to approximate \(\mathrm{softmax}(QK^\top)\approx \Phi(Q)\Phi(K)^\top\), achieving true \(O(N)\) time and memory with bounded error. When applied to ViT backbones, these methods handle larger images with much lower memory cost.

\paragraph{All-MLP Architectures:}  

MLP-Mixer \cite{tolstikhin2021mlp} differs from both CNNs and ViTs by alternating \emph{token-mixing} MLPs (mixing across \(N\) spatial tokens) and \emph{channel-mixing} MLPs (mixing across \(C\) channels). 
This yields per-layer complexity \(O(NC)\) instead of \(O(N^2)\), and achieves ~84\% top-1 
on ImageNet-1K (with ImageNet-21k pretraining), demonstrating that dense MLPs can approximate spatial interactions effectively.

\paragraph{Pyramid/Hierarchical ViTs:}  
Pyramid Vision Transformer (PVT) \cite{wang2021pyramid} builds a multi-scale pyramid by progressively downsampling tokens: early stages operate on high-resolution grids (\(\tfrac{H}{4}\times\tfrac{W}{4}\)), and deeper stages use “patch merging” to halve spatial dimensions at each level. Within each stage, \emph{Spatial-Reduction Attention (SRA)} pools keys/values by a factor \(r\), reducing sequence length from \(N\) to \(N/r^2\) and complexity to \(O(N\cdot N/r^2)\). PVT matches CNN backbones in detection and segmentation.

Swin Transformer \cite{liu2021swin,liu2022swinV2} introduces \emph{window-based MSA} over non-overlapping \(M\times M\) patches (e.g., \(7\times7\)), reducing complexity to \(O\bigl(\tfrac{N}{M^2}\times M^4\bigr)\). Each stage ends with a \emph{patch merging} layer that concatenates \(2\times2\) tokens and projects them, halving resolution and doubling channels. Crucially, Swin alternates “standard” and “shifted” window partitions: shifted windows (offset by \(\lfloor M/2\rfloor\)) overlap adjacent regions, enabling cross-window context without global attention. Swin-B attains 87.3\% top-1 on ImageNet-1K, with near-linear inference latency.

\paragraph{Distilled and Compact ViTs:}
TinyViT \cite{wu2022tinyvit} uses \emph{pretraining-stage distillation} from a large teacher (e.g., Swin-B/L trained on ImageNet-$21$k). By caching teacher logits and applying neural architecture search under FLOPs/parameter constraints, TinyViT produces 11M–21M parameter models that achieve ~$84.8$–$86.5\%$ top-1 on ImageNet-1K—close to much larger ViTs.

\paragraph{Data-Efficient Image Transformers (DeiT)} 

\cite{touvron2021going} add a learnable \emph{distillation token} that learns from a CNN teacher’s soft logits (e.g., ResNet-$50$) while training on ImageNet-1K alone. Combined with aggressive augmentation (RandAugment, Mixup, CutMix) and regularization (Label Smoothing, Stochastic Depth), DeiT-Small (22M) reaches $83.1\%$ top-1 (vs.\ $77.9\%$ for vanilla ViT), and DeiT-Base ($86$M) hits $85.2\%$ in under three GPU-days, matching ResNet-$152$. Later work~\cite{touvron2022deit} adds self-supervised distillation and token pruning for further efficiency.

Collectively, these efforts—linear-attention, MLP-only designs, hierarchical token pyramids, window-based local attention, and distillation—have greatly extended ViT applicability across resource-constrained tasks. However, the inherent hierarchical structure of images remains only partially integrated into existing attention mechanisms, potentially hindering the overall performance.

\paragraph{Multiscale neural architectures.}
    Several transformer architectures have been proposed in the one-dimensional setting of Natural Language Processing (NLP) that are closely related to the multiscale principles underlying our method.
    
    \textbf{H-Transformer-1D}~\cite{H-transformer1D} introduces a hierarchical attention scheme that restricts full attention to local windows while allowing global information to flow through a tree-like structure. 
    
    \textbf{MRA-Attention}~\cite{MRA-attention} leverages a multiresolution decomposition of attention weights using wavelet transforms to capture both coarse and fine-scale dependencies. 
    
    \textbf{FMMformer}~\cite{FMMformer} builds on the Fast Multipole Method (FMM) to hierarchically group tokens and reduce attention complexity by summarizing distant interactions. 
    
    \textbf{Fast Multipole Attention (FMA)}~\cite{fastmultipoleattention} similarly applies FMM-inspired grouping but in a more generalizable attention framework. 
    
    \textbf{ERWIN}~\cite{zhdanov2025erwin} proposes a multilevel window-based transformer with recursive interpolation between coarse and fine spatial scales in the setting of graph attention.

\subsection{Neural Operators}

The challenge in solving PDEs is the computational burden of conventional numerical methods.  To improve the tractability, a recent line of research investigates how machine learning and especially artificial neural networks can provide efficient surrogate models. A first kind of  approache assumes  the knowledge of the underlying PDE, like PINNs \cite{PINNS_raissi,lu2021learning,review_pinns}. With this knowledge, the neural network is optimized by solving the PDE, which can be considered as a kind of unsupervised learning. However, the difficult optimization process requires tailored training schemes with many iterations~\cite{Krishnapriyan21,Ryck24}.  In a "semi-supervised" way, the recent approach of \citet{Boudec25} recasts the problem as a \textit{learning to learn} task, leveraging either, the PDE and  simulations or observations data. While this method obtained promising results, its memory footprint may limit its large scale usage.  
In this work, we focus  neural operators, which learn directly the solution operator  from data \cite{lu2019deeponet,FNO}. In this line of work, the challenge lies in the model architecture rather than in the optimization process and different kind of models were recently proposed. 

\paragraph{Transformer neural operators} In \cite{fourier_galerkin_transformer} the classical transformer 
was adapted for the first time to operator learning problems related to PDEs. The paper explores two variants, based on Fourier transform and Galerkin method. 
The latter one uses a simplified attention based operator, without softmax normalization. This solutions 
shares the linear complexity with our work but not the same expressivity. Still in the simplfyiing trend, LOCA (Learning Operators with Coupled Attention) \cite{coupled_attention} maps the
input functions to a finite set of features and attends to them by output query location. 

Based on kernel theory, \citet{coupled_attention} introduces an efficient transformer for the operator learning setting was proposed  based on kernel theory.  Recently in \cite{calvello2024continuum} was proposed an interesting way to see attention in the continuos setting and in particular the continuum patched attention. In Universal Physics Transformer~\cite{upt} framework for efficient scaling was proposed based on a coarsoning of the input mesh. In \cite{wang2024cvit} the Continuous vision transformer was proposed as a operator-learning version of the more classical ViT.

In the context of operator learning and graph-structured data, the \textbf{Multipole Graph Neural Operator (MGNO)}~\cite{mgno} extends multipole ideas to irregular domains via message-passing on graph hierarchies. 
Finally, \textbf{V-MGNO}, \textbf{F-MGNO}, and \textbf{W-MGNO}~\cite{migus2022multi} propose variations of MGNO to improve stability.

These works highlight the growing interest in multiscale and hierarchical schemes to improve efficiency and generalization, both in sequence modeling and operator learning. Our work builds on this line by proposing a spatially structured multipole attention mechanism adapted to vision and physical simulation tasks.

Our model is explicitly designed to function as a neural operator~\cite{survey_kovachki}. To qualify as a neural operator, a model must satisfy the following key properties. 
First, it should be capable of handling inputs and outputs across arbitrary spatial resolutions. Second, it should exhibit discretization convergence — that is, as the discretization of the input becomes finer, the model’s predictions should converge to the true underlying operator governing the physical system. This pose a new challenge to the computer vision community, namely not just learn an image to image function but the underlying operator independently of the resolution. This field saw its first proof of concept with \citet{lu2019deeponet}, who leveraged a universal approximation theorem for nonlinear operators and paved the way for numerous extensions. Fourier Neural operators \cite{FNO} rely on a translation-equivariant kernel and discretize the problem via a global convolution performed computed by a discrete Fourier transform. Building on this foundation, the Wavelet Neural Operator (WNO) \cite{tripura2022wavelet} introduces wavelet-based multiscale localization, enabling kernels that simultaneously capture global structures and fine-grained details. The Multiwavelet Neural Operator (MWNO) \cite{gupta2021multiwavelet} further extends this approach by incorporating multiple resolution components, leading to improved convergence with respect to discretization.

\section{Detailed hyperparameters}

\subsection{Architecture Hyperparameters for Image classification}
Table~\ref{tab:MANO_img_class_hyperparams} summarizes the architectural and training hyperparameters used in our model. Below, we provide brief comments on each of them.
he first block in Table~\ref{tab:MANO_img_class_hyperparams} corresponds to the standard configuration of the pretrained SwinV2-Tiny model, which we adopt as our backbone.

\begin{itemize}
    \item \textbf{Patch size:} Size of non-overlapping image patches. A value of $4$ corresponds to $4 \times 4$ patches.
    \item \textbf{Input channels:} Number of input channels, set to $3$ for RGB images.
    \item \textbf{Embedding dimension (\texttt{embed\_dim}):} Dimensionality of the token embeddings, controlling model capacity.
    \item \textbf{Global pooling:} Global average pooling is used instead of a [CLS] token at the output.
    \item \textbf{Depths (layers per stage):} Number of transformer blocks in each of the four hierarchical stages, e.g., $[2, 2, 6, 2]$.
    \item \textbf{Number of heads (per stage):} Number of attention heads per stage; increases with depth to maintain representation power.
    \item \textbf{Window size:} Local attention is applied in windows of size $8 \times 8$.
    \item \textbf{MLP ratio:} Ratio between the hidden dimension in the feed-forward MLP and the embedding dimension (e.g., $4.0 \times 96 = 384$).
    \item \textbf{QKV bias:} Whether learnable biases are used in the query/key/value projections (set to \texttt{True}).
    \item \textbf{Dropout rates (\texttt{drop\_rate}, \texttt{proj\_drop\_rate}, \texttt{attn.drop\_rate}):} All standard dropout components are disabled (set to $0$).
    \item \textbf{Drop-path rate (\texttt{drop\_path\_rate}):} Stochastic depth with rate $0.2$ applied to residual connections for regularization.
    \item \textbf{Activation layer:} GELU is used as the non-linearity in MLP layers.
    \item \textbf{Normalization layer:} Layer normalization is applied throughout the network.
    \item \textbf{Pretrained window sizes:} Set to $[0, 0, 0, 0]$ as no pretrained relative position biases are used.
    \item \textbf{Attention sampling rate:} The input to the attention mechanism is downsampled by a factor of $2$, allowing for increased expressivity without a relevant additional computational cost.
    \item \textbf{Attention down-sampling:} A convolutional layer with kernel size $2$ and stride $2$ is used to downsample features between the levels of the multipole attention.
    \item \textbf{Attention up-sampling:} Transposed convolution (kernel size $2$, stride $2$) is used to upsample the features after the windowed attention at each hierarchical level.
    \item \textbf{Number of levels:} Specifies the number of multipole attention levels used at each stage. We found it beneficial to use the maximum number of levels permitted by the spatial resolution.
\end{itemize}

\begin{table}[ht]
  \centering
  \resizebox{\columnwidth}{!}{
  \begin{tabular}{@{} l l @{}}
    \toprule
    \textbf{Hyperparameter}          & \textbf{Value}            \\ 
    \midrule
    \texttt{Patch size}                       & 4                         \\
    \texttt{Input channels}                   & 3                         \\
    \texttt{Embedding dimension} (\texttt{embed\_dim})   & 96                        \\
    \texttt{Global pooling}                   & \texttt{avg}              \\
    \texttt{Depths} (layers per stage)        & [2,\,2,\,6,\,2]           \\
    \texttt{Number of heads} (per stage)      & [3,\,6,\,12,\,24]         \\
    \texttt{Window size}                     & 8                         \\
    \texttt{MLP ratio}                        & 4.0                       \\
    \texttt{qkv bias} (boolean)               & \texttt{True}             \\
    \texttt{Dropout rate} (\texttt{drop\_rate})         & 0.0                       \\
    \texttt{Projection‐drop rate} (\texttt{proj\_drop\_rate}) & 0.0                       \\
    \texttt{Attention‐drop rate} (\texttt{attn\_drop\_rate})  & 0.0                       \\
    \texttt{Drop‐path rate} (\texttt{drop\_path\_rate})    & 0.2                       \\
    \texttt{Activation layer}                 & \texttt{gelu}             \\
    \texttt{Normalization layer} (flag)       & \texttt{True}             \\
    \texttt{Pretrained window sizes}          & [0,\,0,\,0,\,0]           \\
    \midrule
    \texttt{Attention sampling rate}          & 2                         \\
    \texttt{Attention down-sampling}          & \texttt{conv}             \\
    \texttt{kernel size}          & 2             \\
    \texttt{stride}          & 2             \\
    \texttt{Attention up-sampling}            & \texttt{conv transpose}   \\
    \texttt{kernel size}          & 2             \\
    \texttt{stride}          & 2             \\
    \texttt{number of levels}                & [3,\,2,\,1,\,1]           \\
    \bottomrule
  \end{tabular}
  }
  \caption{MANO Hyperparameters for image classification}
  \label{tab:MANO_img_class_hyperparams}
\end{table}

\subsection{Architecture Hyperparameters for Darcy Flow}

Table~\ref{tab:MANO_darcy_hyperparams} reports the main architectural hyperparameters used in our MANO model for solving the Darcy flow problem. Below, we provide a brief description of each.

\begin{itemize}
    \item \textbf{channels}: Number of input channels; set to $3$ because we concatenate the two spatial coordinate with the permeability coefficient.
    \item \textbf{patch size}: Patch size used to partition the input grid; set to $1$ to retain full spatial resolution, ideal for dense prediction tasks.
    \item \textbf{domain dim}: Dimensionality of the input domain; set to $2$ for 2D PDEs like Darcy flow.
    \item \textbf{stack regular grid}: Indicates whether the input discretization is regular and should be stacked; set to \texttt{true}.
    
    \item \textbf{dim}: Embedding dimension of the token representations.
    \item \textbf{dim head}: Dimensionality of each individual attention head.
    \item \textbf{mlp dim}: Hidden dimension of the MLP layers following attention.
    \item \textbf{depth}: Total number of transformer blocks.
    \item \textbf{heads}: Number of self-attention heads in each attention block.
    \item \textbf{emb dropout}: Dropout rate applied to the input embeddings.

    \item \textbf{Attention sampling rate:} The input to the attention mechanism is downsampled by a factor of $2$, allowing for increased expressivity without a relevant additional computational cost.
    \item \textbf{Attention down-sampling:} A convolutional layer with kernel size $2$ and stride $1$ is used to downsample features between the levels of the multipole attention.
    \item \textbf{Attention up-sampling:} Transposed convolution (kernel size $2$, stride $1$) is used to upsample the features after the windowed attention at each hierarchical level.
    \item \textbf{att dropout}: Dropout rate applied within the attention block.

    \item \textbf{Window size:} Local attention is applied in windows of size $2 \times 2$.
    \item \textbf{local attention stride}: Stride with which local windows are applied; controls overlap in attention.

    \item \textbf{positional encoding}: Whether explicit positional encodings are added; set to \texttt{false} in our setting.
    \item \textbf{learnable pe}: Whether the positional encoding is learnable; also disabled here.
    \item \textbf{pos enc coeff}: Scaling coefficient for positional encodings, if used; \texttt{null} since not applicable.
\end{itemize}
\begin{table}[ht]
  \centering
  \resizebox{\columnwidth}{!}{
  \begin{tabular}{@{} l l @{}}
    \toprule
    \textbf{Hyperparameter}          & \textbf{Value}            \\ 
    \midrule
    \texttt{channels} & 3 \\
    \texttt{patch size} & 1 \\
    \texttt{domain dim} & 2 \\
    \texttt{stack regular grid} & true \\
    
    \texttt{dim} & 128 \\
    \texttt{dim head} & 32 \\
    \texttt{mlp dim} & 128 \\
    \texttt{depth} & 8 \\
    \texttt{heads} & 4 \\
    \texttt{emb dropout} & 0.1 \\
    
    \texttt{Attention sampling rate}          & 2                         \\
    \texttt{Attention down-sampling}          & \texttt{conv}             \\
    \texttt{kernel size}          & 2             \\
    \texttt{stride}          & 1             \\
    \texttt{Attention up-sampling}            & \texttt{conv transpose}   \\
    \texttt{kernel size}          & 2             \\
    \texttt{stride}          & 1             \\
    \texttt{att dropout} & 0.1 \\
    \texttt{window size} & 2 \\
    \texttt{local attention stride} & 1 \\
    
    \texttt{positional encoding} & false \\
    \texttt{learnable pe} & false \\
    \texttt{pos enc coeff} & null \\
    
    \bottomrule
  \end{tabular}}
  \caption{MANO Hyperparameters for Darcy flow}
  \label{tab:MANO_darcy_hyperparams}
\end{table}

\section{Implementation details}

All our experiments are implemented in PyTorch.

\subsection{Model checkpoints}
Our experiments in image classification use the following pre-trained models from HuggingFace on ImageNet\cite{imagenet}:
\begin{itemize}
    \item ViT-base available at \url{https://huggingface.co/google/vit-base-patch16-224}
    \item DeiT-small available at \url{https://huggingface.co/facebook/deit-small-patch16-224}
    \item SwinV2 available at \url{https://huggingface.co/timm/swinv2_tiny_window8_256.ms_in1k}
\end{itemize}
We initialize our MANO model by loading the full weights of the pretrained SwinV2-Tiny.

\section{Data Augmentation}
During training, in the case of image classification, we apply standard data augmentations to improve generalization. Specifically, the training pipeline includes:
\begin{itemize}
    \item \texttt{Resize} to a fixed resolution, matching the input size expected by the pretrained models;
    \item \texttt{RandomCrop} with a crop size equal to the resized resolution, using a padding of $4$ pixels;
    \item \texttt{RandomHorizontalFlip};
    \item \texttt{ToTensor} conversion;
    \item \texttt{Normalize} using dataset-specific mean and standard deviation statistics.
\end{itemize}

At test time, images are resized (if necessary), converted to tensors, and normalized using the same statistics as in training.

For numerical simulations, we do not apply any data augmentation.